\documentclass[sigconf]{acmart}
\AtBeginDocument{%
  \providecommand\BibTeX{{%
    \normalfont B\kern-0.5em{\scshape i\kern-0.25em b}\kern-0.8em\TeX}}}

\setcopyright{acmcopyright}
\copyrightyear{2022}
\acmYear{2022}

\acmConference[MM '22]{the 30th ACM
International Conference on Multimedia}{October 10--14, 2022}{Lisbon, Portugal}
\acmPrice{15.00}
\acmISBN{978-1-4503-XXXX-X/18/06}

\acmSubmissionID{2007}

\usepackage{pifont}
\usepackage{color}
\usepackage{subfigure}
\usepackage{url}
\usepackage{enumitem}
\usepackage{algorithm}
\usepackage{algorithmic}
\def\etal{\emph{et al.}}

\begin{document}

\title{RSTAM: An Effective Black-Box Impersonation Attack on Face Recognition using a Mobile and Compact Printer}


\author{Xiaoliang Liu}
\affiliation{%
 \institution{Nanjing University}
 \city{Nanjing}
 \country{China}}

\author{Furao Shen}
\affiliation{%
 \institution{Nanjing University}
 \city{Nanjing}
 \country{China}}

\author{Jian Zhao}
\affiliation{%
 \institution{Nanjing University}
 \city{Nanjing}
 \country{China}}

\author{Changhai Nie}
\affiliation{%
 \institution{Nanjing University}
 \city{Nanjing}
 \country{China}}

\renewcommand{\shortauthors}{XXXXX and XXXXX, et al.}

\begin{abstract}
 Face recognition has achieved considerable progress in recent years thanks to the development of deep neural networks, but it has recently been discovered that deep neural networks are vulnerable to adversarial examples. This means that face recognition models or systems based on deep neural networks are also susceptible to adversarial examples. However, the existing methods of attacking face recognition models or systems with adversarial examples can effectively complete white-box attacks but not black-box impersonation attacks, physical attacks, or convenient attacks, particularly on commercial face recognition systems. In this paper, we propose a new method to attack face recognition models or systems called RSTAM, which enables an effective black-box impersonation attack using an adversarial mask printed by a mobile and compact printer. First, RSTAM enhances the transferability of the adversarial masks through our proposed random similarity transformation strategy. Furthermore, we propose a random meta-optimization strategy for ensembling several pre-trained face models to generate more general adversarial masks. Finally, we conduct experiments on the CelebA-HQ, LFW, Makeup Transfer (MT), and CASIA-FaceV5 datasets. The performance of the attacks is also evaluated on state-of-the-art commercial face recognition systems: Face++, Baidu, Aliyun, Tencent, and Microsoft. Extensive experiments show that RSTAM can effectively perform black-box impersonation attacks on face recognition models or systems.
 \end{abstract}




\maketitle

\section{Introduction}
Face recognition has advanced significantly in recent years with the development of deep neural networks~\cite{parkhi2015deep,schroff2015facenet,liu2017sphereface,wang2018cosface,deng2019arcface}. In our daily lives, face recognition has been used in a wide range of applications due to the excellent performance currently available. These applications include security-sensitive applications such as mobile phone unlocking and payment, door locks, airport and railway station check-in, financial industry authentication, and other similar applications. Unfortunately, many studies~\cite{goodfellow2014explaining,madry2018towards,carlini2017towards,dong2018boosting} have found that deep neural networks are vulnerable to adversarial examples. Unsurprisingly, face recognition based on deep neural networks is also vulnerable to adversarial examples~\cite{dong2019efficient,zhong2020towards,yang2021attacks}. 

Dong~\etal~\cite{dong2019efficient}, Zhong~\etal~\cite{zhong2020towards}, and Yang~\cite{yang2021attacks} generate adversarial examples with strong perturbations by decision-based attack, gradient-based attack, and generative adversarial network (GAN)~\cite{goodfellow2014generative}, respectively, which can effectively perform digital adversarial attacks on face recognition. However, implementing these global perturbations in the physical world is unrealistic. Therefore, they cannot be employed in actual attacks. In most cases, these methods are used to evaluate the security of face recognition rather than to carry out actual attacks. Recently, many methods for physical attacks on face recognition have also been proposed~\cite{sharif2016accessorize,sharif2019general,nguyen2020adversarial,komkov2021advhat,yin2021adv}. Sharif~\etal~\cite{sharif2016accessorize,sharif2019general} and Komkov~\etal~\cite{komkov2021advhat} are perfect for white-box physical attacks, but they are hard to perform black-box attacks on face recognition models or systems. Nevertheless, realistic face recognition environments are commonly in a black-box state. Nguyen~\etal~\cite{nguyen2020adversarial} used a projector to perform adversarial light projection physical attacks, but such attacks are not convenient in reality. Yin~\etal~\cite{yin2021adv} can implement transferable physical attacks on face recognition, but physical attacks on commercial face recognition systems are not ideal. Similarly, Yin~\etal~\cite{yin2021adv} perform poorly when confronted with low-quality face images.

We summarize that an effective adversarial attack method on face recognition should be capable of conducting black-box attacks, physical attacks, impersonation attacks, convenient attacks, attacks on low-quality target images, and attacks against commercial systems. Previous studies have shown that it is not easy to effectively implement black-box physical impersonation attacks on face recognition. To address these challenges simultaneously, we propose an effective black-box impersonation attack method on face recognition, called RSTAM, which implements a physical attack employing an adversarial mask printed by a mobile and compact printer. To begin, we design an initial binary mask, as shown in Figure~\ref{fig:RSTAM}. Secondly, we propose a random similarity transformation strategy, which can enhance the diversity of the inputs and thus the transferability of the adversarial masks. Following that, we propose a random meta-optimization strategy for ensembling several pre-trained face models to generate more transferable adversarial masks. Finally, we perform experiments on two high-resolution face datasets CelebA-HQ~\cite{karras2018progressive}, Makeup Transfer (MT)~\cite{li2018beautygan}, and two low-quality face datasets LFW~\cite{huang2008labeled}, CASIA-FaceV5~\cite{casiafacev5}. Meanwhile, we perform evaluations on five state-of-the-art commercial face recognition systems: Face++~\cite{faceplusplus}, Baidu~\cite{baidu}, Aliyun~\cite{aliyun}, Tencent~\cite{tencent} and Microsoft~\cite{microsoft}. Our experiments show that our proposed method RSTAM can effectively perform black-box impersonation attacks on commercial face recognition systems and low-pixel target images. Moreover, RSTAM can implement convenient physical attacks through the use of a Canon SELPHY CP1300~\cite{cp1300}, a mobile and compact printer. The main contributions of our work are summarized as follows.
\begin{itemize}[leftmargin=*]
\setlength{\itemsep}{0pt}
\setlength{\parskip}{0pt}
\setlength{\parsep}{0pt}
	\item We design an initial binary mask for the adversarial masks.
	\item We propose a random similarity transformation strategy that can improve the transferability of the adversarial masks by increasing the diversity of the inputs. Moreover, we use only one hyperparameter to control the random similarity transformation with four degrees of freedom (4DoF). 
	\item We propose a random meta-optimization strategy to perform an ensemble attack using several pre-trained face models. This strategy enables us to extract more common gradient features from the face models, thereby increasing the transferability of the adversarial masks.
	\item Experiments demonstrate that RSTAM is an effective attack method on face recognition capable of performing black-box attacks, physical attacks, convenience attacks, impersonation attacks, attacks on low-quality images, and attacks against state-of-the-art commercial face recognition systems.
\end{itemize}

\section{Related Works}
\subsection{Adversarial Attacks}
Adversarial attacks fall into two broad categories: white-box attacks and black-box attacks. For white-box attack methods, they have full access to the target model and and the majority of them are gradient-based attacks, such as Fast Gradient Sign Method (FGSM)~\cite{goodfellow2014explaining}, Projected Gradient Descent (PGD)~\cite{madry2018towards}, and Carlini \& Wagner's method (C\&W)~\cite{carlini2017towards}. FGSM is a single-step gradient-based attack method that shows that linear features of deep neural networks in high-dimensional space are sufficient to generate adversarial examples. PGD is a multi-step extension of the FGSM attack that generates more powerful adversarial examples for white-box attacks. C\&W is an optimization-based attack method that also happens to be a gradient-based attack method. However, in practice, what exists is more of a black-box scenario. White-box attack methods tend to lack transferability and fail to attack target models with unknown parameters and gradients. Therefore, more researchers have focused on black-box attack methods. 

Black-box attack methods can be classified into three categories: transfer-based, score-based, and decision-based. The transfer-based attacks generate the adversarial examples with a source model and then transfer them to a target model to complete the attack, in which we do not need to know any parameters or gradients of the target model. MI-FGSM~\cite{dong2018boosting} suggests incorporating momentum into the attack process to stabilize the update direction and increase the transferability of the generated adversarial examples. DI$^2$-FGSM~\cite{xie2019improving} first proposes improving the transferability of the adversarial examples by increasing the diversity of the inputs. The translation-invariant attack method (TI-FGSM)~\cite{dong2019evading} and the affine-invariant attack method (AI-FGSM) ~\cite{xiang2021improving} are used to further improve the transferability and robustness of adversarial examples. Score-based attacks can simply know the output score of the target model and estimate the gradient of the target model by querying the output score~\cite{ilyas2018black,cheng2019improving,li2019nattack}. Decision-based attacks assume that the attack is performed in a more challenging situation where only the output labels of the classifier are known. Boundry Attack~\cite{brendel2018decision} and Evolutionary Attack~\cite{dong2019efficient} are effective methods for dealing with this attack setting. 

\subsection{Adversarial Attacks on Face Recognition}
Adversarial attacks on face recognition come in two common forms: dodging attacks and impersonation attacks. The purpose of dodging attacks is to reduce the confidence of the similarity of the same identity pair in order to evade face recognition. Impersonation attacks attempt to fool face recognition by using one identity to mimic another. Both technically and practically, impersonation attacks are more challenging and practical than dodging attacks. As a result, we concentrate on impersonation attack methods. Dong~\etal ~\cite{dong2019efficient} proposed a decision-based adversarial attack on face recognition. Zhong~\etal~\cite{zhong2020towards} increased the diversity of agent face recognition models by using the dropout~\cite{srivastava2014dropout} technique to improve the transferability of the adversarial examples. Yang~\etal~\cite{yang2021attacks} introduced a GAN~\cite{goodfellow2014generative} to generate adversarial examples for impersonation attacks on face recognition. Dong~\etal ~\cite{dong2019efficient}, Zhong~\etal~\cite{zhong2020towards}, and Yang~\etal~\cite{yang2021attacks} are the digital-based attacks on face recognition, which makes them hard to implement in the real world. At this point, there are also many researchers who have proposed methods for physical-based attacks on face recognition. Sharif~\etal~\cite{sharif2016accessorize,sharif2019general} proposed a way to perform real-physical attacks on face recognition by printing out a pair of eyeglass frames. Komkov~\etal~\cite{komkov2021advhat} proposed a physical attack method that can print an adversarial sticker using a color printer and put it on a hat to complete the attack. Nguyen~\etal~\cite{nguyen2020adversarial} proposed an adversarial light projection attack method that uses a projector for physical attack. Yin~\etal~\cite{yin2021adv} generated eye makeup patches by GAN and printed them out and stuck them around the eyes to perform physical attacks. Methods \cite{sharif2016accessorize,sharif2019general,komkov2021advhat} are perfect for white-box physical attacks. But realistic environments are often black-box situations and they are hard to attack with black-box face recognition models or systems. Although the method~\cite{yin2021adv} can perform a transferable black-box attack, it is ineffective for low-pixel face images and for commercial face recognition systems. 

Compared with the previous methods, we propose an adversarial attack method on face recognition, RSTAM, which can effectively accomplish the black-box impersonation attack, both on low-pixel face pictures and on commercial face recognition systems. Furthermore, RSTAM can carry out a physical attack with the help of a mobile and compact printer.

\begin{figure*}[t]
  \centering
  \includegraphics[width=1.0\linewidth]{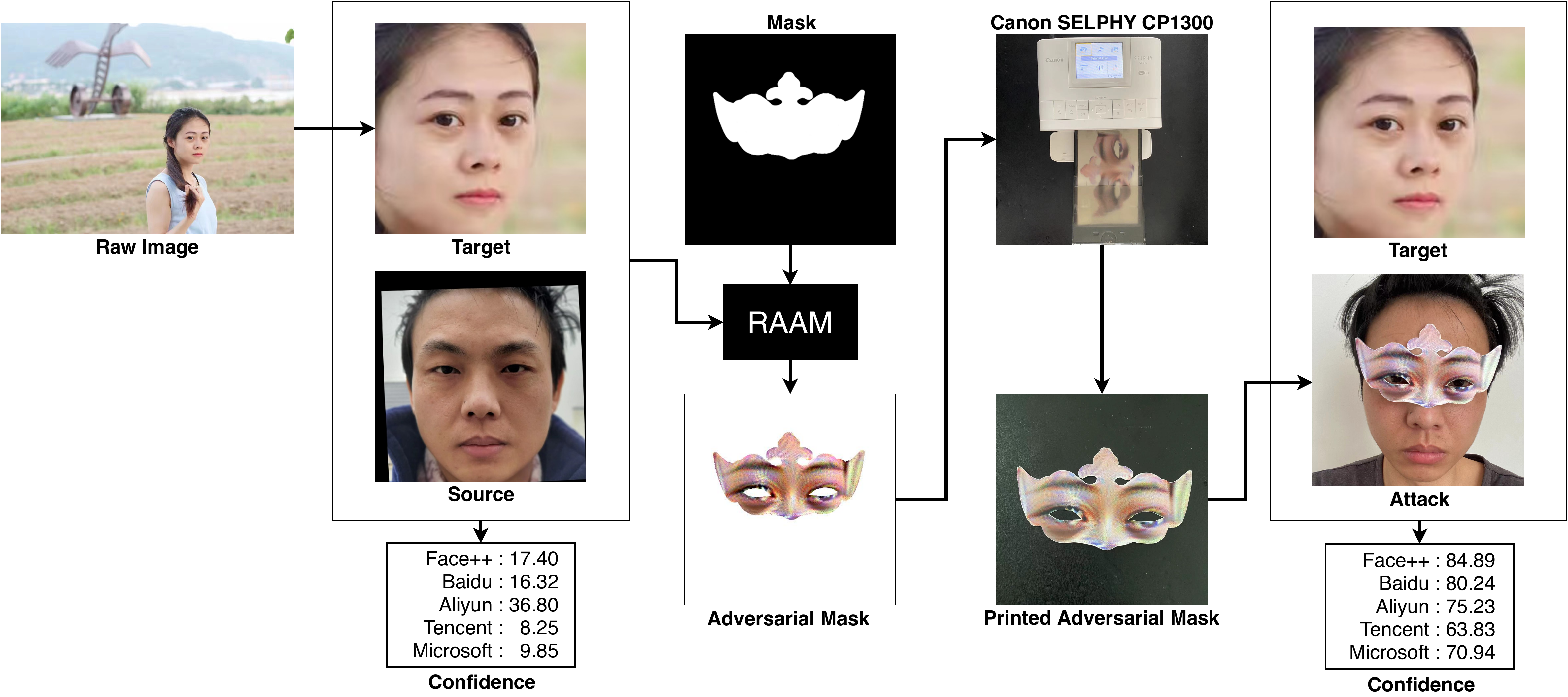}
  \caption{An example of RSTAM. The raw image of the target identity is from a social network. The initial binary mask is designed by us. We can observe that the confidence scores between the attack and the target are significantly greater than those between the source and the target.}
  \label{fig:RSTAM}
  \Description{An example of RSTAM attack.}
\end{figure*}
\section{Methodology}
\subsection{Overview}

Figure~\ref{fig:RSTAM} shows an example of the RSTAM. The raw image of the target identity is from a social network. Firstly, we apply the facial landmark detection method~\cite{JLS21} to generate corresponding facial landmarks of the raw image. Then we obtain the aligned target image according to the facial landmarks. Similarly, we can use the facial landmarks in the source image to obtain the eye region of the attacker. After that, we use our proposed method RSTAM to generate an adversarial mask and print out the adversarial mask using the Canon SELPHY CP1300~\cite{cp1300}, which is a mobile and compact printer. Finally, the attacker with the printed adversarial mask is obtained. We can see from Figure~\ref{fig:RSTAM} that the similarity confidence between the attack and the target is high on all five commercial face recognition systems, with Face++~\cite{faceplusplus} 84.89\%, Baidu~\cite{baidu} 80.24\%, Aliyun~\cite{aliyun} 75.23\%, Tencent~\cite{tencent} 63.83\%, and Microsoft~\cite{microsoft} 70.94\%. Figure~\ref{fig:RSTAM} further shows that we can effectively perform physical impersonation attacks on the commercial face recognition systems using a photo of the target identity on a social network.

\subsection{Adversarial Mask}
In this section, we will give a detailed description for the adversarial mask. Let $\mathbf{x}^{t}$ denote a face image of the target identity, $\mathbf{x}^{s}$ denote a source image of the attacker, $\mathbf{x}^{adv}$ denote an attack image of the attacker with an adversarial mask and $f(\mathbf{x}) : \mathbf{X} \rightarrow \mathbb{R}^d$ denote a face recognition model that extracts a normalized feature representation vector for an input image $\mathbf{x} \in \mathbf{X} \subset \mathbb{R}^n$. Our goal for the aversarial mask attack is to solve the following constrained optimization problem,
\begin{equation}
\begin{aligned}
&\mathop{argmin} \limits_{\mathbf{x}^{adv}} \mathcal{L}(f(\mathbf{x}^{adv}),f(\mathbf{x}^{t})), \\
	&s.t. \ \mathbf{x}^{adv} \odot (1-\mathbf{M}) = \mathbf{x}^s \odot (1-\mathbf{M}),
\end{aligned}
\label{eq:1}
\end{equation}
where $\mathcal{L}$ is a cosine similarity loss function,
\begin{equation}
	\mathcal{L}(\mathbf{v}^s,\mathbf{v}^t) = 1-cos(\mathbf{v}^s,\mathbf{v}^t).
\end{equation}
\begin{figure}[tb]
  \centering
  \includegraphics[width=1.0\linewidth]{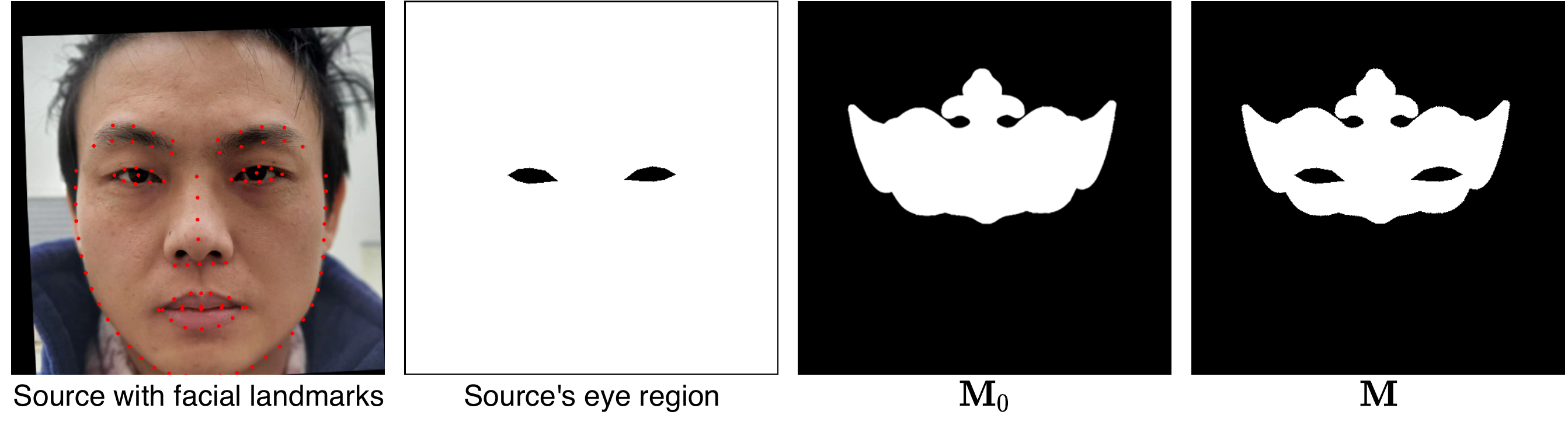}
  \caption{An example of generating a binary mask $\mathbf{M}$ using the initial binary mask $\mathbf{M}_0$ and the facial landmarks of the source.}
  \label{fig:mask}
  \Description{An example of mask.}
\end{figure}
$\odot$ is the element-wise product, and $\mathbf{M}$ is a binary mask. The binary mask $\mathbf{M}$ is used to constrain that only one pixel can be perturbed when its corresponding position is 1 in $\mathbf{M}$. We design an initial binary mask $\mathbf{M}_0$, and then use the initial binary mask $\mathbf{M}_0$ and the facial landmarks of the source image to generate the corresponding binary mask $\mathbf{M}$. Figure~\ref{fig:mask} shows an example of the binary mask $\mathbf{M}$ we generate using the initial binary mask $\mathbf{M}_0$ and the facial landmarks of the source image. 

Let $\mathbf{A}$ denote an adversarial mask. We can generate an adversarial mask $\mathbf{A}$ with the $\ell_\infty$-norm perturbations~\cite{goodfellow2014explaining,madry2018towards} by performing a multi-step update as
\begin{equation}
\begin{aligned}
	 &\mathbf{x}^{adv}_0 = \mathbf{x}^s \odot (1-\mathbf{M}) + \mathbf{x}^t \odot \mathbf{M}, \\	
	&{\mathbf{x}}_{n+1}^{adv} = {Clip}_{\mathbf{x}^{adv}_0,\epsilon}({\mathbf{x}}_{n}^{adv} - \alpha \cdot \mathbf{sign}(\nabla_{{\mathbf{x}}^{adv}_{n}} \mathcal{L}(f({\mathbf{x}}^{adv}_{n}), f({\mathbf{x}}^{t})))\odot \mathbf{M}), \\
	& \mathbf{A} = {\mathbf{x}}_{n+1}^{adv} \odot \mathbf{M},
\end{aligned}
\end{equation}
where $\alpha$ is a perturbation step size, $\mathbf{sign}(\cdot)$ is the sign function, ${Clip}_{\mathbf{x}^{adv}_0,\epsilon}$ denotes element-wise clipping, aiming to restrict $\mathbf{x}^{adv}$ with in the $\ell_\infty$-bound of $\mathbf{x}^{adv}_0$. $\epsilon$ is a perturbation bound, $ \Vert \mathbf{x}^{adv}-\mathbf{x}_{0}^{adv} \Vert_\infty \leq \epsilon$ .

\subsection{Random Similarity Transformation}
The similarity transformation, which has four degrees of freedom (4DoF), consists of translational, rotational, and scaling transformations. It is commonly used for face alignment~\cite{tadmor2016learning,liu2017sphereface}. Many previous studies~\cite{xie2019improving,gao2020patch,dong2019evading,xiang2021improving} have demonstrated that the transferability of adversarial examples can be greatly improved by increasing the diversity of the inputs. In order to improve the transferability of the adversarial masks, we propose a random similarity transformation strategy to increase the diversity of the inputs. Moreover, our strategy requires only one hyperparameter to control the random similarity transformation with 4DoF. Let $U(a,b)$ denote the uniformly distributed random sampling function from $a$ to $b$ and $\beta$ denote a hyperparameter. At each iteration we can obtain a random similarity transformation matrix $\mathbf{T}$ by the following strategy,

\begin{equation}
	\begin{aligned}
		t_x & = U(-\beta W,\beta W),\\
		t_y & = U(-\beta H,\beta H),\\
		\theta & = U(-\beta \pi/2,\beta \pi/2),\\
		s & = U(1-\beta,1+\beta),\\
		\mathbf{T} &=\left[ \begin{array}{ccc}
			1 & 0 & t_x \\
			 0 & 1 & t_y \\
			0 & 0 & 1
		\end{array} \right] \left[ \begin{array}{ccc}
			cos(\theta) & sin(\theta) & 0 \\
			 -sin(\theta) & cos(\theta) & 0 \\
			0 & 0 & 1
		\end{array} \right]\left[ \begin{array}{ccc}
			s & 0 & 0 \\
			 0 & s & 0 \\
			0 & 0 & 1
		\end{array} \right]\\
		 &=\left[ \begin{array}{ccc}
			s \cdot cos(\theta) & s \cdot sin(\theta) & t_x\\
			-s \cdot sin(\theta)  & s \cdot cos(\theta) & t_y\\
			0 & 0 & 1
		\end{array} \right].
	\end{aligned}
\end{equation}
where $W$ and $H$ are the width and height of the input image. Let $(p_x,p_y)$ denote one coordinate of the input image. We can use the similarity transformation matrix $\mathbf{T}$ to obtain the corresponding transformed coordinates $(p_x^t, p_y^t)$,
\begin{equation}
\begin{aligned}
\left[ \begin{array}{c}
			p_x^t  \\
			p_y^t \\
			1
		\end{array} \right]&=\mathbf{T}\left[ \begin{array}{c}
			p_x  \\
			p_y \\
			1
		\end{array} \right].\\
\end{aligned}
\end{equation}
Finally, we generate the transformed input image by bilinear interpolation. 

\subsection{Adversarial Mask with Random Similarity Transformation (RSTAM)}
Xie~\etal~\cite{xie2019improving} first found that the transferability of adversarial examples could be further improved by increasing the diversity of inputs. Methods~\cite{dong2019evading,gao2020patch,xiang2021improving} also demonstrate this finding. From the description in Section 3.2, we can consider that the adversarial masks are a type of the adversarial examples, so we can also use this finding as well. For the adversarial mask attack on face recognition, we propose a random similarity transformation strategy to enhance the diversity of the input face images, which is described in Section 3.3. Using this strategy we propose the adversarial mask attack method with random similarity transformation, RSTAM. Algorithm block~\ref{alg:RSTAM} describes the detailed algorithm for the $\ell_\infty$-bound RSTAM, RSTAM$_\infty$. Similarly, the RSTAM attack can also use the $\ell_2$-norm perturbations, RSTAM$_2$. In RSTAM$_2$, we get $\bar{\mathbf{x}}^{adv}_{n+1}$ by
\begin{equation}
\begin{aligned}
	\bar{\mathbf{x}}^{adv}_{n+1} = {Clip}_{[\mathbf{x}^{adv}_0-\epsilon, \mathbf{x}^{adv}_0+\epsilon]}(\mathbf{x}^{adv}_{n} - \alpha \cdot \frac{\mathbf{g}_n}{\Vert \mathbf{g}_n \Vert_2} \odot \mathbf{M}).
\end{aligned}
\end{equation}

\begin{algorithm}[tb]
\normalsize
	\caption{RSTAM$_\infty$ algorithm}	
	\label{alg:RSTAM}
	\begin{algorithmic}[1]
		\REQUIRE{A face image $\mathbf{x}^t$ of the target identity; A source image $\mathbf{x}^s$ of the attacker; the initial binary mask $\mathbf{M}_0$ of our design; the facial landmarks $lms^{s}$ of the source image; A target face model $f$. }
		\REQUIRE{Iterations $N$; the perturbation bound $\epsilon$; the perturbation step size $\alpha$; a hyperparameter $\beta$ for the random similarity transformation.}
		\REQUIRE{The binary mask generation function $\boldsymbol{GenM}$; the random similarity transformation function $\boldsymbol{RST}$; the sign function $\boldsymbol{sign}$.}
		\REQUIRE{The cosine similarity loss function $\mathcal{L}$.}
			\STATE $\mathbf{M} = \boldsymbol{GenM}(\mathbf{M}_0,lms^{s})$
			\STATE $\mathbf{x}^{adv}_0 = \mathbf{x}^s \odot (1-\mathbf{M}) + \mathbf{x}^t \odot \mathbf{M}$
			\FOR{$n=0$ to $N-1$}
				\STATE $\mathbf{g}_n = \nabla_{\mathbf{x}^{adv}_n}{\mathcal{L}(f(\boldsymbol{RST}(\mathbf{x}^{adv}_n,\beta)),f(\mathbf{x}^t))}$
				\STATE $\bar{\mathbf{x}}^{adv}_{n+1} = \mathop{Clip}_{[\mathbf{x}^{adv}_0-\epsilon, \mathbf{x}^{adv}_0+\epsilon]}(\mathbf{x}^{adv}_{n} - \alpha \cdot \boldsymbol{sign}(\mathbf{g}_n) \odot \mathbf{M})$
				\STATE $\mathbf{x}^{adv}_{n+1} = \mathop{Clip}_{[\mathbf{0},\mathbf{1}]}(\bar{\mathbf{x}}^{adv}_{n+1})$
			\ENDFOR
			\STATE $\mathbf{A}_N = \mathbf{x}^{adv}_{N} \odot \mathbf{M}$
		\RETURN $\mathbf{x}^{adv}_{N}, \mathbf{A}_N$
	\end{algorithmic}
\end{algorithm}

When multiple pre-trained face models are available, it is natural to consider using the ensemble idea to obtain more transferable adversarial examples. Ensemble methods are also often used in research and competitions to improve performance and robustness~\cite{dietterich2000ensemble,seni2010ensemble}. Dong~\etal~\cite{dong2018boosting} demonstrated that the transferability of adversarial examples can be effectively improved by applying ensemble methods. However, the number of pre-trained models available is limited. Therefore, Dong's hard ensemble method is still prone to ``overfitting'' pre-trained face models. Meta-learning has been proposed as a framework to address the challenging few-shot learning setting~\cite{finn2017model,sun2019meta,jamal2019task}. Inspired by meta-learning, we propose a random meta-optimization strategy for ensembling several pre-trained face models to generate adversarial masks. Different from meta-learning, which is used to update the parameters of the neural network model, the random meta-optimization strategy assumes the network model as data and then updates the adversarial masks directly. Algorithm block~\ref{alg:RSTAM_meta} describes in detail the RAAM$_{\infty}^{meta}$ algorithm for performing ensemble attacks using the random meta-optimization strategy.

\begin{algorithm}[tb]
\normalsize
	\caption{RSTAM$_{\infty}^{meta}$ algorithm}	
	\label{alg:RSTAM_meta}
	\begin{algorithmic}[1]
		\REQUIRE{A face image $\mathbf{x}^t$ of the target identity; A source image $\mathbf{x}^s$ of the attacker; the initial binary mask $\mathbf{M}_0$ of our design; the facial landmarks $lms^{s}$ of the source image; target face models $F=[f_1,f_2,...,f_m]$.}
		\REQUIRE{Iterations $N$; the perturbation bound $\epsilon$; a perturbation step size $\alpha$; a hyperparameter $\beta$ for the random similarity transformation.}
		\REQUIRE{The binary mask generation function $\boldsymbol{GenM}$; the random similarity transformation function $\boldsymbol{RST}$; the sign function $\boldsymbol{sign}$.}
		\REQUIRE{The cosine similarity loss function $\mathcal{L}$.}
			\STATE $\mathbf{M} = \boldsymbol{GenM}(\mathbf{M}_0,lms^{s})$
			\STATE $\mathbf{x}^{adv}_0 = \mathbf{x}^s \odot (1-\mathbf{M}) + \mathbf{x}^t \odot \mathbf{M}$
			\FOR{$n=0$ to $N-1$}
				\STATE $f_{que}^{meta} = Random.choice(F)$, a model is randomly selected from F as the meta-query model.
				\STATE $F_{sup}^{meta} = F.remove(f_{que}^{meta})$, the remaining models in F are used as meta-support models.
				\STATE $\mathbf{g}_{que}^{meta}=\mathbf{0}$
				\STATE $\mathbf{g}_{sup}^{meta}=\mathbf{0}$
				\FOR{$f_{sup}^{meta}$ in $F_{sup}^{meta}$}
				\STATE	$\mathbf{g}_{sup} = \nabla_{\mathbf{x}^{adv}_n}{\mathcal{L}(f_{sup}^{meta}(\boldsymbol{RST}(\mathbf{x}^{adv}_n,\beta)),f_{sup}^{meta}(\mathbf{x}^t))}$
				\STATE $\bar{\mathbf{x}}^{meta} = \mathop{Clip}_{[\mathbf{x}^{adv}_0-\epsilon, \mathbf{x}^{adv}_0+\epsilon]}(\mathbf{x}^{adv}_{n} - \alpha \cdot \boldsymbol{sign}(\mathbf{g}_{sup}) \odot \mathbf{M})$
				\STATE $\mathbf{x}^{meta} = \mathop{Clip}_{[\mathbf{0},\mathbf{1}]}(\bar{\mathbf{x}}^{meta})$
				
				\STATE $\mathbf{g}_{que} =  \nabla_{\mathbf{x}^{meta}}{\mathcal{L}(f_{test}^m(\boldsymbol{RA}(\mathbf{x}^{meta},\beta)),f_{test}^{meta}(\mathbf{x}^t))}$
				\STATE $\mathbf{g}_{que}^{meta}= \mathbf{g}_{que}^{meta} + \mathbf{g}_{que}$
				\STATE $\mathbf{g}_{sup}^{meta} = \mathbf{g}_{sup}^{meta} + \mathbf{g}_{sup}$
				\ENDFOR
				\STATE $\mathbf{g}_n = \frac{1}{m-1}(\mathbf{g}_{sup}^{meta}+\mathbf{g}_{que}^{meta})$
				\STATE $\bar{\mathbf{x}}^{adv}_{n+1} = \mathop{Clip}_{[\mathbf{x}^{adv}_0-\epsilon, \mathbf{x}^{adv}_0+\epsilon]}(\mathbf{x}^{adv}_{n} - \alpha \cdot \boldsymbol{sign}(\mathbf{g}_n) \odot \mathbf{M})$
				\STATE $\mathbf{x}^{adv}_{n+1} = \mathop{Clip}_{[\mathbf{0},\mathbf{1}]}(\bar{\mathbf{x}}^{adv}_{n+1})$
			\ENDFOR
			\STATE $\mathbf{A}_N = \mathbf{x}^{adv}_{N} \odot \mathbf{M}$
		\RETURN $\mathbf{x}^{adv}_{N}, \mathbf{A}_N$
	\end{algorithmic}
\end{algorithm}

\section{Experiments}
\subsection{Experimental Setup}
 \noindent\textbf{Datasets.} In the experiments, we use four public face datasets, which contain two high-resolution face datasets CelebA-HQ~\cite{karras2018progressive}, Makeup Transfer (MT)~\cite{li2018beautygan}, and two low -quality face datasets LFW~\cite{huang2008labeled}, CASIA-FaceV5~\cite{casiafacev5}.
\begin{itemize}[leftmargin=*]
\setlength{\itemsep}{0pt}
\setlength{\parskip}{0pt}
\setlength{\parsep}{0pt}
\item CelebA-HQ is a high-quality version of CelebA~\cite{liu2015faceattributes} that consists of 30,000 images at $1024\times1024$ resolution. 
 
\item LFW is made up of 13,233 low-quality facial images gathered from the internet. There are 5749 identities in this collection, with 1680 people having two or more photos. 

\item MT is a facial makeup dataset that includes 3834 female face images, with 2719 makeup images and 1115 non-makeup images. 
  
\item CASIA-FaceV5 contains 2500 facial images of 500 subjects from Asia and all face images are captured using Logitech USB camera. 
\end{itemize}

In order to evaluate the performance of the attacks, we randomly select 500 different identity pairs from the CelebA-HQ and LFW datasets, respectively. Furthermore, we randomly select 1000 different identity makeup images from the MT dataset to make up 500 different identity pairs, and 500 subjects of CASIA-FaceV5 are randomly paired into 250 different identity pairs for the experiment.
~\\[4pt]
\noindent\textbf{Face Recognition Models and Face Recognition Systems.} In our experiments, we use five face recognition models, FaceNet~\cite{schroff2015facenet}, MobileFace, IRSE50, IRSE101, and IR151~\cite{deng2019arcface}, and five commercial face recognition systems, Face++~\cite{faceplusplus}, Baidu~\cite{baidu}, Aliyun~\cite{aliyun}, Tencent~\cite{tencent} and Microsoft~\cite{microsoft}. Because the API of Aliyun's face recognition system is not available to individual users, we only use the web application of Aliyun's face recognition system for our experiments. In Microsoft's face recognition system, we use ``recognition\_04'' face recognition model and ``detection\_03'' face detection model, which are the latest versions of Microsoft's face recognition system. All other face recognition systems use the default version.
~\\[4pt]
\noindent\textbf{Evaluate Metrics.} 
For impersonation attacks on face recognition models, the attack success rate ($ASR$)~\cite{deb2020advfaces,zhong2020towards,yin2021adv} is reported as an evaluation metric,
\begin{equation}
	\begin{aligned}
		ASR=\frac{\sum_{i=1}^N 1_{\tau} (cos[f(x^t_i), f({x^{adv}_i})] > \tau)}{N} \times 100\%
	\end{aligned}
\end{equation}
where $N$ is the number of pairs in the face dataset, $1_{\tau}$ denotes the indicator function, $\tau$ is a pre-determined threshold. For each victim facial recognition model, $\tau$ will be determined at 0.1\% False Acceptance Rate ($FAR$) on all possible image pairs in LFW, i.e. FaceNet 0.409, MobileFace 0.302, IRSE50 0.241, and IR152 0.167. 

For the evaluation of attacks on face recognition systems, we report the mean confidence scores ($MCS$) on each dataset as an evaluation metric,  
\begin{equation}
	MCS = \frac{\sum^N_{i=1} \emph{conf}_i }{N} \times 100\%
\end{equation}
where \emph{conf} is a confidence score between the target and the attack returned from the face recognition system API, $N$ is the number of pairs in the face dataset.
~\\[4pt]
\noindent\textbf{Implementation Details.} We first resize all the input images from all datasets to $512 \times 512$ and normalize to $[0,1]$. The following is the setting of our main comparison method in the experiment.
\begin{itemize}[leftmargin=*]
\setlength{\itemsep}{0pt}
\setlength{\parskip}{0pt}
\setlength{\parsep}{0pt}
\item  PASTE is the standard baseline that we use to directly paste the target image's associated binary mask $M$ region onto the attacker's source image and then perform the impersonation attack.
\item  AM$_{\infty}$ is the adversarial mask attack method with the $\ell_\infty$-bound. The perturbation step size $\alpha$ is set to 0.003. The number of iterations $N$ is set to 2000. the perturbation bound $\epsilon$ is set to 0.3. The target face model $f$ uses IRSE101 and the remaining models, FaceNet, MobileFace, IRSE50 and IR151, are used as victim models for the black box attack.
\item  RSTAM$_{\infty}$ is the $\ell_\infty$-bound RSTAM. The hyperparameter $\beta$ is set to 0.2. Other settings are the same as on AM$_{\infty}$.
\item  RSTAM$_{\infty}^{all}$ uses all five face recognition models as target face models without using the random meta-optimization ensemble strategy. Assume that the target face models $F=[f_1,f_2,f_3,f_4,f_5]$, we can get the $n_{th}$ update gradient as
\begin{equation}
	\mathbf{g}_n = \frac{1}{5}\sum^{5}_{i=1} \nabla_{\mathbf{x}^{adv}_n}{\mathcal{L}(f_i(\boldsymbol{RST}(\mathbf{x}^{adv}_n,\beta)),f_i(\mathbf{x}^t))}.
\end{equation}
Other settings are the same as on RSTAM$_{\infty}$.
\item  RSTAM$_{\infty}^{meta}$ is the $\ell_\infty$-bound RSTAM with using the random meta-optimization ensemble strategy. The target models $F$ use all five face recognition models. Other settings are the same as on RSTAM$_{\infty}$. 
\item  RSTAM$_{2}$ is the $\ell_2$-bound RSTAM. The perturbation step size $\alpha$ is set to 2. Other settings are the same as on RSTAM$_{\infty}$.
\item  RSTAM$_{2}^{meta}$ is the $\ell_2$-bound RSTAM with using the random meta-optimization ensemble strategy. The perturbation step size $\alpha$ is set to 2. Other settings are the same as on RSTAM$_{\infty}^{meta}$.
\end{itemize}
Additionally, we implement our codes based on the open source deep learning platform PyTorch~\cite{paszke2019pytorch}.
\subsection{Digital-Environment Experiments}
In this section, we will present the outcomes of the black-box impersonation attack in the digital world. Firstly, we will report quantitative results and qualitative results in the datasets CelebA-HQ, LFW, MT, and CASIA-FaceV5, respectively. Next, we will report the results of the hyperparameter $\beta$ sensitivity experiments.
~\\[4pt]
\noindent\textbf{Quantitative Results.} Table 1-4 show the quantitative results on digital images. Table~\ref{tab:celeba} shows the results of the black-box impersonation attack on the CelebA-HQ dataset, which is a high-definition multi-attribute face dataset including annotations with 40 attributes per image. The images in this collection span a wide range of position variants as well as background clutter. We can see from Table~\ref{tab:celeba} that the $ASR$ of RSTAM$_\infty$ for Face Models is much higher than PASTE benchmark and AM$_\infty$, 64.20\% (RSTAM$_\infty$) vs. 27.40\% (PASTE) vs. 31.60\% (AM$_\infty$) on FaceNet, 63.20\% vs. 48.00\% vs. 43.80\% on MobileFace, 91.00\% vs. 54.80\% vs. 59.60\% on IRSE50, and 91.80\% vs. 41.20\% vs. 51.20\% on IR151. The hard ensemble attack method RSTAM$^{all}_\infty$ increases the $MCS$ on Face++ (74.24\% vs. 71.94\%) and Baidu (72.54\% vs. 70.80\%) but decreases on Tencent (49.50\% vs. 50.63\%) and Microsoft (49.12\% vs. 53.97\%) for the Face Systems in Table~\ref{tab:celeba}. Thus, we can think that the hard ensemble attack method is not the best ensemble attack method, and the ensemble attack method RSTAM$^{meta}_{\infty}$ based on our proposed random meta-optimization strategy can further improve the performance of the ensemble attack method, 74.76\% (RSTAM$^{meta}_{\infty}$)  vs. 74.24\% (RSTAM$^{all}_\infty$) on Face++, 72.83\% vs. 72.54\% on Baidu, 50.88\% vs. 49.50 on Tencent, and 50.58\% vs. 49.12\% on Microsoft. Lastly, we can also see in Table~\ref{tab:celeba} that the $\ell_2$-bound RSTAM has better attack performance than the $\ell_\infty$-bound in the black-box attack on commercial face recognition system.

Table~\ref{tab:lfw} provides the results of the black-box impersonation attack on the low-quality face dataset LFW.  From Table~\ref{tab:lfw}, we can observe that the RSTAM attack is still effective on low-quality face images. The $MCS$ of RSTAM$^{meta}_2$ on LFW can reach 70.29\% in Face++, 70.08\% in Baidu, 51.45\% in Tencent and 50.13\% in Microsoft.

 Table~\ref{tab:mt} presents the results of the black box impersonation attack on the female makeup face images of the MT dataset and Table~\ref{tab:casia} presents the results of the attack on the Asian face dataset CASIA-FaceV5. Compared with the multi-attribute datasets CelebA-HQ and LFW, the Face Models show lower robustness under relatively single-attribute face datasets MT and CASIA, and PASTE can then achieve a higher $ASR$. In contrast, commercial face recognition systems show similar robustness to single-attribute face datasets and multi-attribute face datasets. It can also be demonstrated that RSTAM, our proposed black-box impersonation attack method, can effectively work with single-attribute or multi-attribute datasets, high-quality or low-quality images, face recognition models, and commercial face recognition systems.
\begin{table}[htbp]
\centering
\caption{The results of digital black-box impersonation attacks on the CelebA-HQ dataset. The attack evaluation metric for face models uses $ASR$ (\%), while the attack evaluation metric for face systems uses $MCS$ (\%). The highlighted values represent the best in each column.}
\label{tab:celeba}
\scalebox{0.65}{
\begin{tabular}{l|cccc|cccc}
\toprule
&\multicolumn{4}{c|}{Face Models} & \multicolumn{4}{c}{Face Systems}\\
 & FaceNet &MobileFace & IRSE50 & IR151 & Face++ & Baidu & Tencent & Microsoft  \\ \midrule
 PASTE & 27.40 & 48.00 & 54.80 & 41.20 & 66.21 &  61.98 & 30.37 & 29.60 \\ \midrule
 AM$_{\infty}$& 31.60 & 43.80 & 59.60 & 51.20 & 65.66  & 61.44 & 33.42  & 32.87 \\
 RSTAM$_{\infty}$ & \textbf{64.20}  & \textbf{63.20}  & \textbf{91.00}& \textbf{91.80} & 71.94 & 70.80 & 50.63 & 53.97  \\ 
 RSTAM$_{\infty}^{all}$ & - & - & - & - & 74.24 & 72.54  & 49.50 & 49.12  \\
 RSTAM$_{\infty}^{meta}$ & - & - & - & - & 74.76 & 72.83 & 50.88 & 50.58  \\ \midrule
 RSTAM$_{2}$ & 60.80 & 62.4 & 89.4 & \textbf{91.80} & 71.18 & 70.75 & \textbf{52.60} &\textbf{55.67}  \\ 
 RSTAM$_{2}^{meta}$ & - & - & - & - & \textbf{74.80} & \textbf{72.90} & 51.94 & 52.19  \\  
   \bottomrule
\end{tabular}
}
\end{table}

\begin{table}[htbp]
\centering
\caption{The results of digital black-box impersonation attacks on the LFW dataset. The attack evaluation metric for face models uses $ASR$ (\%), while the attack evaluation metric for face systems uses $MCS$ (\%). The highlighted values represent the best in each column.}
\label{tab:lfw}
\scalebox{0.65}{
\begin{tabular}{l|cccc|cccc}
\toprule
&\multicolumn{4}{c|}{Face Models} & \multicolumn{4}{c}{Face Systems}\\
 & FaceNet &MobileFace & IRSE50 & IR151 & Face++ & Baidu & Tencent & Microsoft  \\ \midrule
 PASTE & 24.00 & 36.40 & 44.20 & 21.20 & 58.90 & 51.86  & 27.70 & 22.13  \\ \midrule
 AM$_{\infty}$& 28.00 &  36.60 & 49.40 & 33.60 & 59.65 & 52.85 & 31.04 & 26.37 \\
 RSTAM$_{\infty}$ & \textbf{59.20} & 50.80 & \textbf{85.40} & \textbf{89.00} & 66.11 & 66.94 & 47.34 & 48.88 \\ 
 RSTAM$_{\infty}^{all}$ & - & - & - & - & 69.59 & 69.45 & 48.20 & 45.78  \\
 RSTAM$_{\infty}^{meta}$ & - & - & - & - & 70.00 & 69.82 & 49.45  & 47.71  \\ \midrule
 RSTAM$_{2}$ & 57.60 & \textbf{52.00} & 84.80 & \textbf{89.00} & 65.56 & 66.93 & 51.16 & \textbf{53.18}  \\ 
 RSTAM$_{2}^{meta}$ & - & - & - & - & \textbf{70.29} & \textbf{70.08} & \textbf{51.45} & 50.13 \\ 
   \bottomrule
\end{tabular}
}
\end{table}

\begin{table}[htbp]
\centering
\caption{The results of digital black-box impersonation attacks on the MT dataset. The attack evaluation metric for face models uses $ASR$ (\%), while the attack evaluation metric for face systems uses $MCS$ (\%). The highlighted values represent the best in each column.}
\label{tab:mt}
\scalebox{0.65}{
\begin{tabular}{l|cccc|cccc}
\toprule
&\multicolumn{4}{c|}{Face Models} & \multicolumn{4}{c}{Face Systems}\\
 & FaceNet &MobileFace & IRSE50 & IR151 & Face++ & Baidu & Tencent & Microsoft  \\ \midrule
 PASTE & 86.40 & 61.80 & 71.30  & 42.00 & 63.87 & 55.98 & 28.15 & 28.16  \\ \midrule
 AM$_{\infty}$& 87.00 & 61.20 & 77.40 &  54.80 & 63.62 & 57.07  & 31.26 & 31.41 \\
 RSTAM$_{\infty}$ & \textbf{94.40} & 79.00 & \textbf{95.60} & \textbf{92.20} & 71.26 & 65.81 & 46.43 & 49.06 \\ 
 RSTAM$_{\infty}^{all}$ & - & - & - & - & 72.59 & 67.39 & 45.37 & 47.74  \\
 RSTAM$_{\infty}^{meta}$& - & - & - & - & 73.06 & 67.98  & 46.08  & 48.90 \\ \midrule
 RSTAM$_{2}$ & 93.80 & \textbf{79.40} & 94.20 & \textbf{92.20} & 70.73 & 66.17 & \textbf{48.23} & \textbf{51.84}  \\ 
 RSTAM$_{2}^{meta}$ & - & - & - & - & \textbf{73.12}  & \textbf{68.18}  & 47.44  & 50.65  \\  
  \bottomrule
\end{tabular}
}
\end{table}

\begin{table}[htbp]
\centering
\caption{The results of digital black-box impersonation attacks on the CASIA-FaceV5 dataset. The attack evaluation metric for face models uses $ASR$ (\%), while the attack evaluation metric for face systems uses $MCS$ (\%). The highlighted values represent the best in each column.}
\label{tab:casia}
\scalebox{0.65}{
\begin{tabular}{l|cccc|cccc}
\toprule
&\multicolumn{4}{c|}{Face Models} & \multicolumn{4}{c}{Face Systems}\\
 & FaceNet &MobileFace & IRSE50 & IR151 & Face++ & Baidu & Tencent & Microsoft  \\ \midrule
 PASTE & 92.40 & 80.80 & 89.20 & 66.40 & 59.53 & 52.02 & 27.99 & 42.82 \\ \midrule
 AM$_{\infty}$& 92.00 & 79.20  & 90.80 & 72.40 & 63.44 & 56.52 & 34.05  & 46.01 \\
 RSTAM$_{\infty}$ & \textbf{97.20} & 87.60 & 98.40 & 97.60 & 71.70 & 68.27 & 42.59 & 57.34  \\ 
 RSTAM$_{\infty}^{all}$ & - & - & - & - & 72.84 & 69.91 & 44.39 & 57.77  \\
 RSTAM$_{\infty}^{meta}$ & - & - & - & - & 73.31 & 70.26 & 44.96  & 58.88 \\ \midrule
 RSTAM$_{2}$ &\textbf{97.20}  & \textbf{90.00} & \textbf{98.80} & \textbf{98.00} & 71.44 & 69.41 & \textbf{47.29} & \textbf{62.62}  \\ 
 RSTAM$_{2}^{meta}$ & - & - & - & - & \textbf{73.82} & \textbf{70.76} & 46.71 & 60.87 \\ 
   \bottomrule
\end{tabular}
}
\end{table}

\begin{figure*}[tbh]
  \centering
  \includegraphics[width=1.0\linewidth]{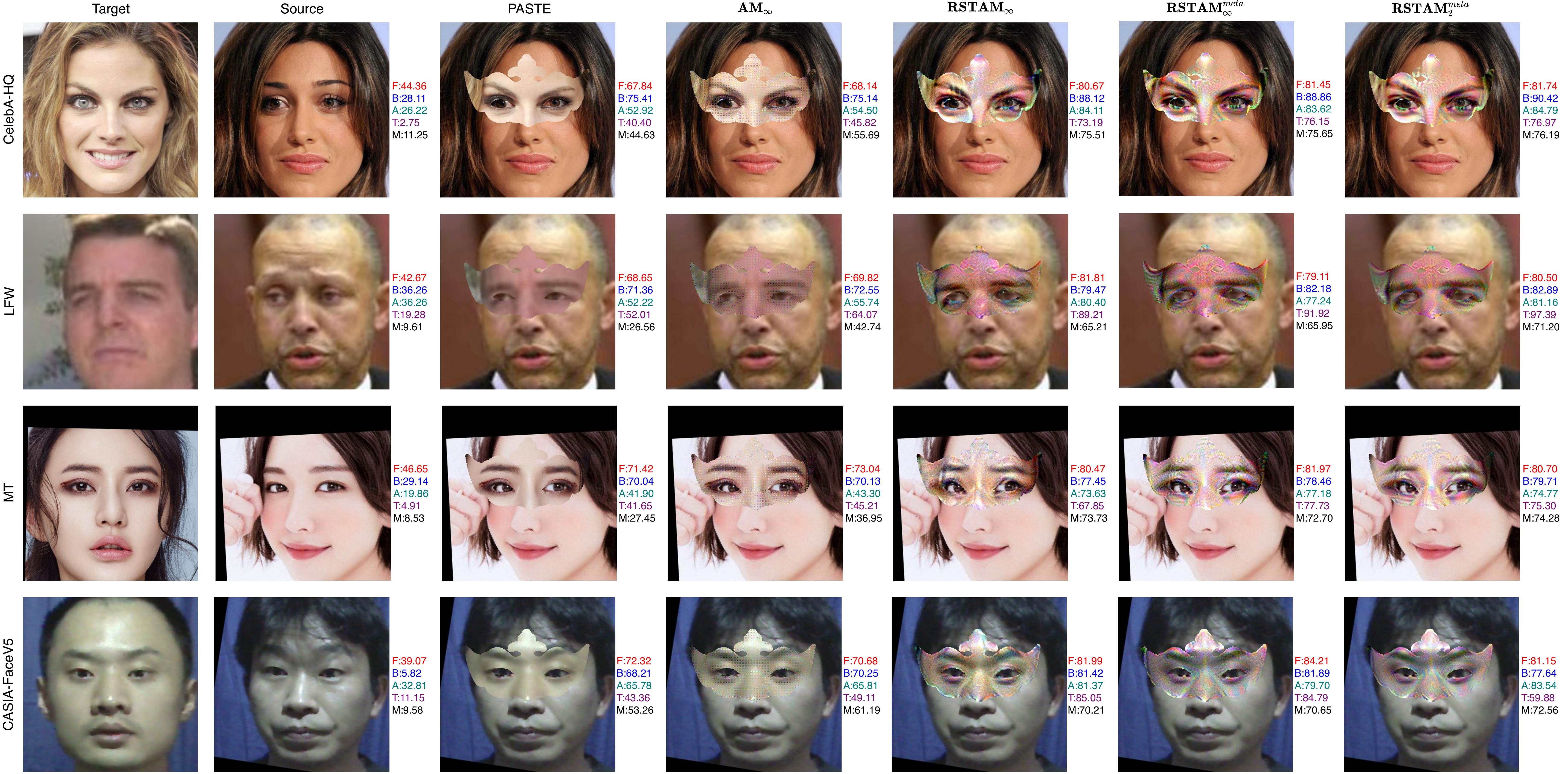}
  \caption{The visualization results of digital black-box impersonation attacks on five commercial face recognition systems. The confidence scores are pasted to the right of each attack, (\textcolor{red}{F:Face++}, \textcolor{blue}{B:Baidu}, \textcolor{teal}{A:Aliyun}, \textcolor{violet}{T:Tencent}, \textcolor{black}{M:Microsoft}).}
  \label{fig:digital-attack}
  \Description{digital-attack}
\end{figure*}
\begin{figure}[htbp]
  \centering
\includegraphics[width=1.0\linewidth]{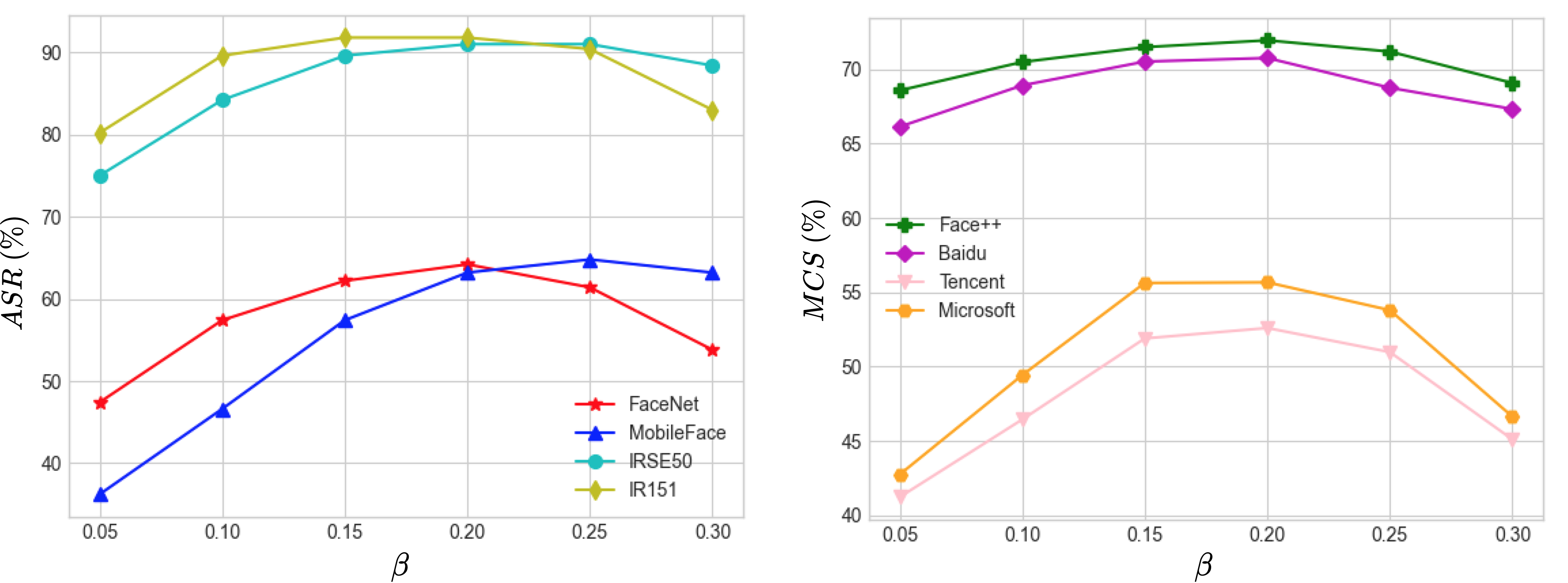}
  \caption{The results of the hyperparameter $\beta$ sensitivity experiments using the RSTAM$_\infty$ attack on the CelebA-HQ dataset.}
  \label{fig:beta}
  \Description{fig:beta}
\end{figure}
\noindent\textbf{Qualitative Results.} The results of the qualitative are shown in Figure~\ref{fig:digital-attack}. As illustrated in Figure 2, the confidence scores between the targets and the attacks generated via RSTAM are much higher than the confidence scores between the targets and the sources. The confidence scores of the attacks using RSTAM are mostly greater than 70\%. In particular, the confidence score between the attack and the target can reach 97.39\% on the LFW using RSTAM$^{meta}_2$ for the Tencent face recognition system. 
~\\[4pt]
\noindent\textbf{Sensitivity of the Hyperparameter $\beta$.}
The hyperparameter $\beta$ is used to control the random similarity transformation with 4DoF, which plays an important role in RSTAM. We perform sensitivity experiments for the hyperparameter $\beta$ using RSTAM$_\infty$ on the CelebA-HQ dataset. The results of the hyperparameter $\beta$ sensitivity experiments are shown in Figure~\ref{fig:beta}. As shown in Figure~\ref{fig:beta}, we suggest that the hyperparameter $\beta$ is set between 0.15 and 0.25. In all experiments except the sensitivity experiment of $\beta$, we set the hyperparameter $\beta$ to 0.2.

\begin{figure}[tb]
  \centering
\includegraphics[width=0.8\linewidth]{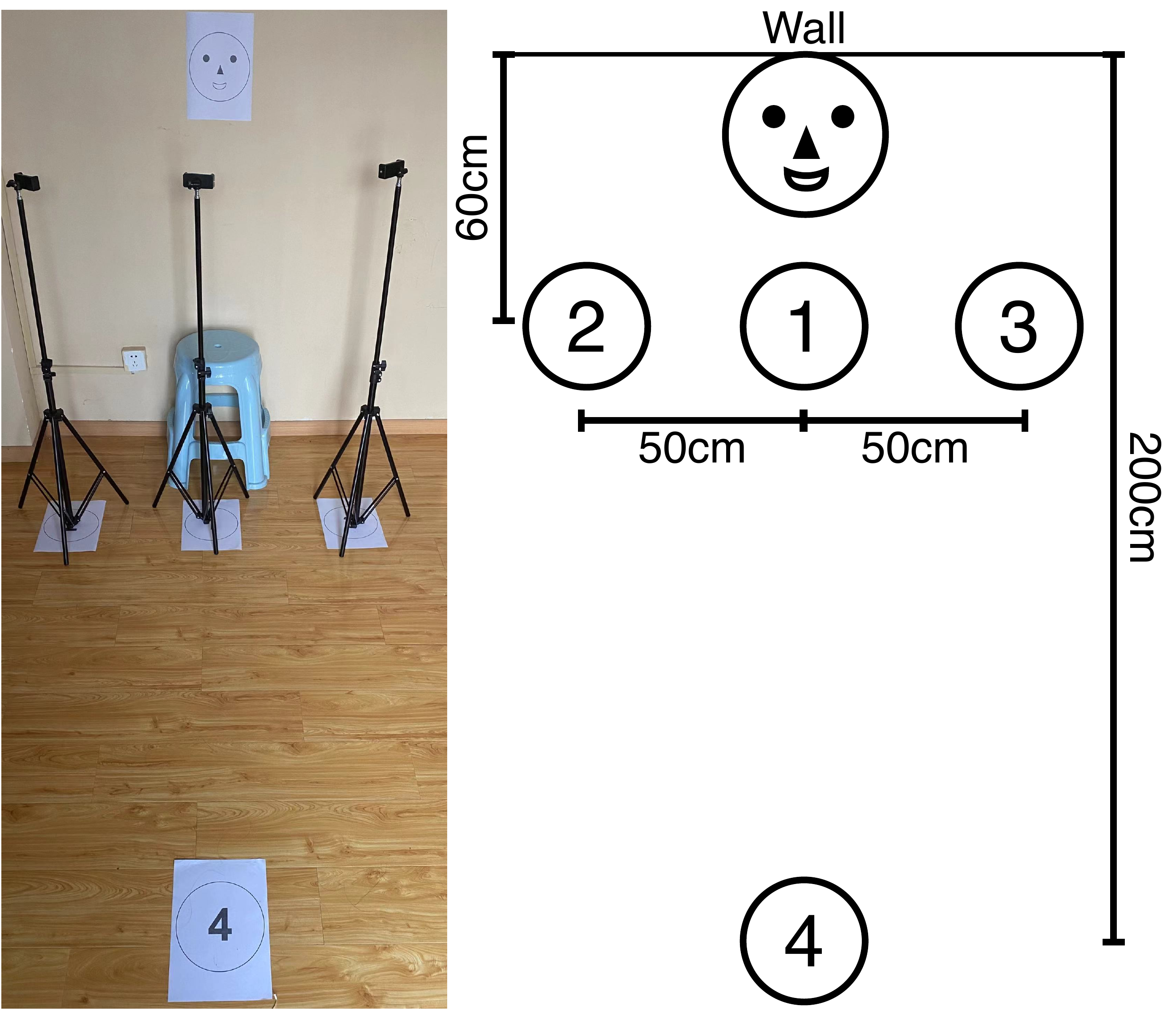}
  \caption{Realistic environmental settings for physical attack experiments.}
  \label{fig:env-setting}
  \Description{env-setting}
\end{figure}

\begin{figure*}[tbph]
  \centering
  \includegraphics[width=0.8\linewidth]{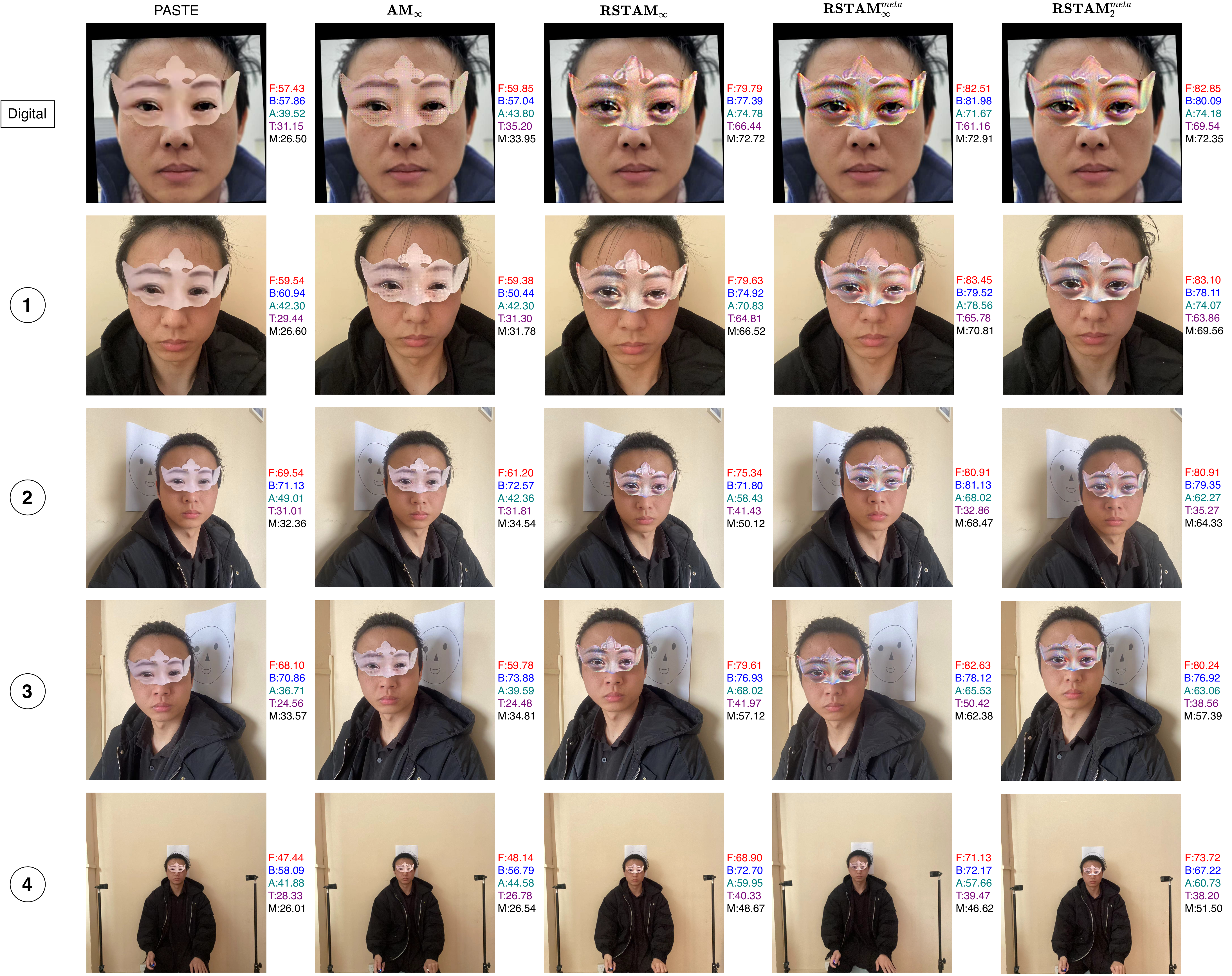}
  \caption{The visualization results of physical black-box impersonation attacks on five commercial face recognition systems. The target and source are the same as in Figure~\ref{fig:RSTAM}. The confidence scores are pasted to the right of each attack, (\textcolor{red}{F:Face++}, \textcolor{blue}{B:Baidu}, \textcolor{teal}{A:Aliyun}, \textcolor{violet}{T:Tencent}, \textcolor{black}{M:Microsoft}). }
  \label{fig:phy-att}
  \Description{phy-att}
\end{figure*}
\subsection{Physical-Realizability Experiments}
The successful completion of an attack in the digital world does not guarantee that it can be applied to the physical world. Moreover, compared to digital attacks, physical attacks have more practical value in the real world. Therefore, we use a mobile and compact printer, Canon SELPHY CP1300, to print out adversarial masks to carry out experiments on physical attacks. The setup of the physical environment is shown in Figure~\ref{fig:env-setting}. For shooting, the camera makes use of the iPhone 11 Pro Max's 12 MP front-facing camera as well as a Bluetooth remote control.

Figure 5 shows the visualization results of our physical black-box impersonation attacks against the five state-of-the-art commercial face recognition systems. The target and source are the same as in Figure~\ref{fig:RSTAM}. Although the printed adversarial masks exhibit distortion in comparison to the digital adversarial masks, the confidence scores of the physical attacks at position \ding{172} are not much reduced, and even increase on RSTAM$^{meta}_{\infty}$ and RSTAM$^{meta}_{2}$ against the Face++. This also shows that RSTAM is effective in using a mobile and compact printer to implement physical attacks in a realistic environment. Except for the attack on the Tencent face system, RSTAM can maintain a high confidence score for commercial face recognition systems at various positions. Moreover, RSTAM can have good attack performance at the long-range position \ding{175}, where the face is of low quality. Similarly, in the physical world, the RSTAM$^{meta}_{\infty}$ and RSTAM$^{meta}_{2}$ ensemble attack methods based on random meta-optimization strategy have superior attack performance.

\section{Conclusions}
In this paper, we propose a black-box impersonation attack method on face recognition, RSTAM. In order to improve the transferability of the adversarial masks, we propose a random similarity transformation strategy for increasing the input diversity and a random meta-optimization strategy for ensembling several pre-trained face models to generate more general adversarial masks. Finally, we perform experimental validation on four public face datasets: CelebA, LFW, MT, CASIA-FaceV5, and five commercial face recognition systems: Face++, Baidu, Aliyun, Tencent, and Microsoft. The experiments adequately demonstrate that RSTAM is an effective attack on face recognition. Furthermore, RSTAM can be easily implemented as a physical black-box impersonation attack using a mobile and compact printer. We can also find that the current commercial face recognition systems are not very secure. We can easily collect real face images of target identities from social networks and then complete the impersonation attacks with RSTAM. Therefore, our further work will focus on how to effectively defend against RSTAM and achieve a more robust and secure face recognition model.
\balance
\bibliographystyle{ACM-Reference-Format}
\bibliography{ref.bib}

\end{document}